\documentclass[11pt,a4paper]{article}
\usepackage{acl}

\usepackage{times}
\usepackage{latexsym}

\usepackage{microtype}

\usepackage{times}
\usepackage{latexsym}
\usepackage{booktabs}
\usepackage[T1]{fontenc}

\usepackage[utf8]{inputenc}

\usepackage{microtype}
\usepackage{amsmath}
\usepackage{booktabs}
\usepackage{array}
\newcolumntype{L}{>{$}l<{$}}
\newcolumntype{C}{>{$}c<{$}}
\newcolumntype{R}{>{$}r<{$}}

\usepackage{times}
\usepackage{latexsym}

\usepackage{microtype}

\usepackage{times}
\usepackage{latexsym}

\usepackage[T1]{fontenc}

\usepackage[utf8]{inputenc}

\usepackage{multirow}
\usepackage{array, tabularx, caption, boldline}
\usepackage{graphicx}
\usepackage{cellspace}
\usepackage{algpseudocode}  
\usepackage{amsmath}
\usepackage{multicol}  
\usepackage{multirow} 
\usepackage{flexisym}
\usepackage{graphicx}  
\usepackage{array}

\usepackage{microtype}

\usepackage{dsfont}

\usepackage{tabularx}
\makeatletter
\def\hlinewd#1{%
\noalign{\ifnum0=`}\fi\hrule \@height #1 %
\futurelet\reserved@a\@xhline}
\makeatother

\usepackage{times}
\usepackage{latexsym}

\usepackage{algpseudocode}  
\usepackage{amsmath}
\usepackage{multicol}  
\usepackage{multirow} 
\usepackage{graphicx}  
\usepackage{array}
\usepackage{arydshln}
\usepackage{booktabs}
\usepackage{textcomp}

\usepackage{amssymb}
\usepackage{pifont}

\usepackage{cleveref}
\crefformat{section}{\S#2#1#3} 
\crefformat{subsection}{\S#2#1#3}
\crefformat{subsubsection}{\S#2#1#3}

\usepackage{microtype}

\usepackage{adjustbox}
\usepackage{multicol}  
\usepackage{multirow} 
\usepackage{amsmath}
\usepackage{amsfonts}
\usepackage{makecell}
\usepackage{algpseudocode} 
\usepackage{verbatim}
\usepackage{mathtools}

\usepackage{booktabs}

\usepackage{makecell}
\usepackage{cleveref}
\usepackage[normalem]{ulem}

\usepackage{times}
\usepackage{latexsym}

\usepackage{footnote}
\makesavenoteenv{tabular}
\makesavenoteenv{table}

\usepackage[hang,flushmargin]{footmisc}

\usepackage[linesnumbered,ruled]{algorithm2e}
\SetAlFnt{\small}
\SetAlCapNameFnt{\normalsize}
\makeatletter
\newcommand{\nosemic}{\renewcommand{\@endalgocfline}{\relax}}
\newcommand{\dosemic}{\renewcommand{\@endalgocfline}{\algocf@endline}}
\let\oldnl\nl
\newcommand{\nonl}{\renewcommand{\nl}{\let\nl\oldnl}}
\makeatother

\usepackage{enumitem}
\usepackage{tablefootnote}

\usepackage{tabularx}
\makeatletter
\def\hlinewd#1{%
\noalign{\ifnum0=`}\fi\hrule \@height #1 %
\futurelet\reserved@a\@xhline}
\makeatother

\usepackage{cleveref}
\crefformat{section}{\S#2#1#3}
\crefformat{subsection}{\S#2#1#3}
\crefformat{subsubsection}{\S#2#1#3}
\crefrangeformat{section}{\S\S#3#1#4 to~#5#2#6}
\crefmultiformat{section}{\S\S#2#1#3}{ and~#2#1#3}{, #2#1#3}{ and~#2#1#3}
\usepackage{refstyle}
\Crefformat{figure}{#2Figure~#1#3}
\Crefmultiformat{figure}{Figures.~#2#1#3}{ and~#2#1#3}{, #2#1#3}{ and~#2#1#3}
\Crefformat{table}{#2Table~#1#3}
\Crefmultiformat{table}{Tables~#2#1#3}{ and~#2#1#3}{, #2#1#3}{ and~#2#1#3}
\Crefformat{appendix}{Appx.~\S#2#1#3}
\crefformat{algorithm}{Alg.~#2#1#3}

\title{TaCL: Improving BERT Pre-training with\\ Token-aware Contrastive Learning}

\author{Yixuan Su$^{\spadesuit}$~\quad Fangyu Liu$^{\spadesuit}$~\quad Zaiqiao Meng$^{\spadesuit}$~\quad Tian Lan$^{\diamondsuit}$~\quad\\\textbf{Lei Shu}$^{\heartsuit}\thanks{~~Work was done prior to joining Amazon.}$\quad   \textbf{Ehsan Shareghi}$^{\clubsuit \spadesuit}$\quad  \textbf{Nigel Collier}$^{\spadesuit}$\\
$^{\spadesuit}$Language Technology Lab, University of Cambridge \ \ \ \ \ $^\diamondsuit$Beijing Institute of Technology\\
$^\heartsuit$Amazon AWS AI \ \ \ \ \ $^{\clubsuit}$ Department of Data Science and AI, Monash University\\
{\tt \{ys484,fl399,zm324,nhc30\}@cam.ac.uk, lantiangmftby@gmail.com}\\ 
{\tt leishu@amazon.com, ehsan.shareghi@monash.edu}\\
}

\date{}

\begin{document}
\maketitle

\begin{abstract}
Masked language models (MLMs) such as BERT have revolutionized the field of Natural Language Understanding in the past few years. However, existing pre-trained MLMs often output an anisotropic distribution of token representations that occupies a narrow subset of the entire representation space. Such token representations are not ideal, especially for tasks that demand discriminative semantic meanings of distinct tokens. In this work, we propose \textbf{TaCL} (\textbf{T}oken-\textbf{a}ware \textbf{C}ontrastive \textbf{L}earning), a novel continual pre-training approach that encourages BERT to learn an isotropic and discriminative distribution of token representations. TaCL is fully unsupervised and requires no additional data. We extensively test our approach on a wide range of English and Chinese benchmarks. The results show that TaCL brings consistent and notable improvements over the original BERT model. Furthermore, we conduct detailed analysis to reveal the merits and inner-workings of our approach.\footnote{Our code and pre-trained models are publicly available at \url{https://github.com/yxuansu/TaCL}}
\end{abstract}

\section{Introduction}
Since the rising of BERT \citep{DBLP:conf/naacl/DevlinCLT19}, masked language models (MLMs) have become the de facto backbone for almost all natural language understanding (NLU) tasks. Despite their clear success, many existing language models pre-trained with MLM objective suffer from the \textit{anisotropic problem}~\cite{DBLP:conf/emnlp/Ethayarajh19}. That is, their token representations reside in a narrow subset of the representation space, therefore being less discriminative and less powerful in capturing the semantic differences of distinct tokens.
\begin{figure}[tb] 
  \centering    
  \setlength{\abovecaptionskip}{3pt}
  \includegraphics[width=0.43\textwidth]{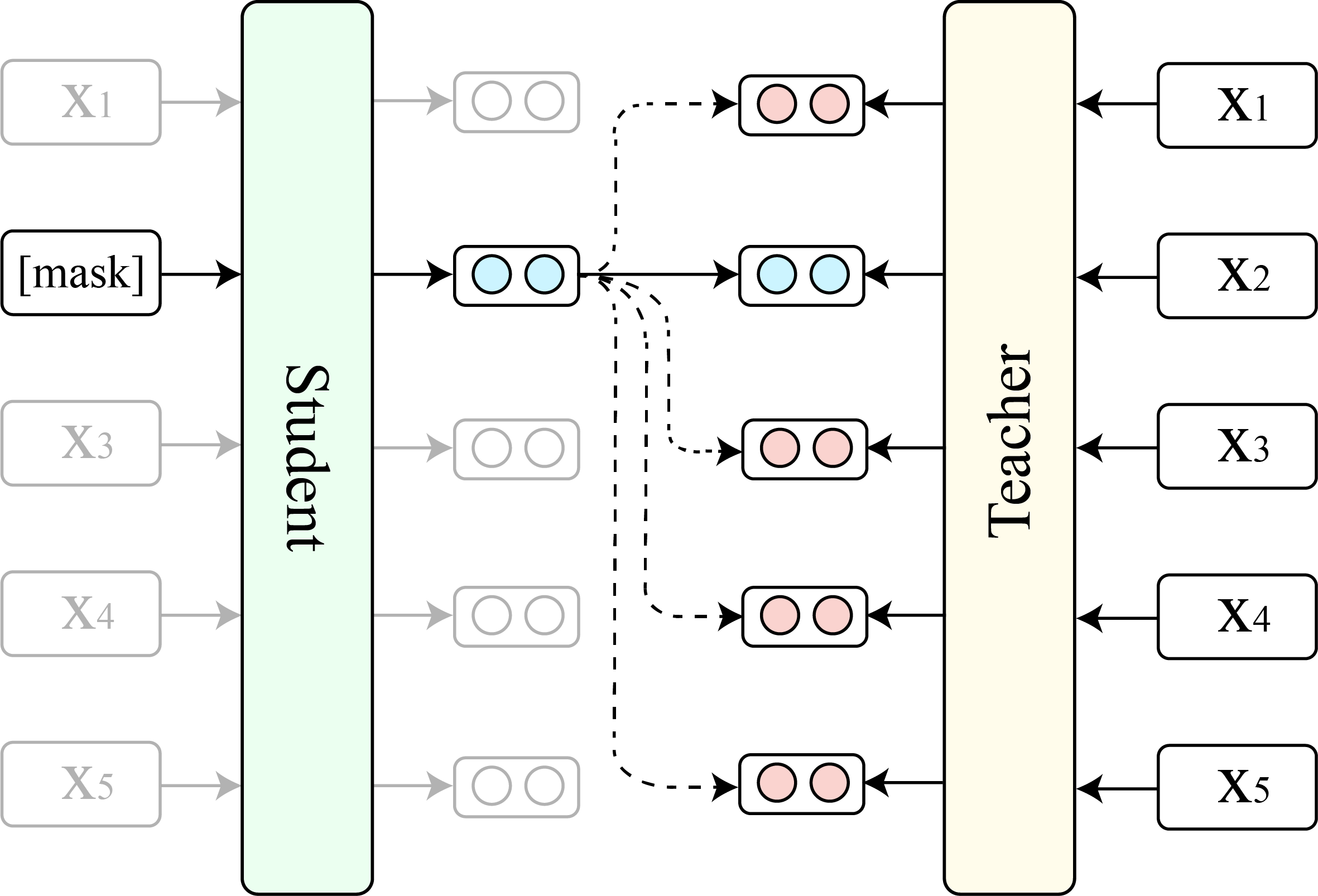}
  \caption{An overview of TaCL. The student learns to make the representation of a masked token closer to its ``reference'' representation produced by the teacher~(solid arrow) and away from the representations of other tokens in the same sequence~(dashed arrows).}
  \label{fig:overview}
  \vspace{-1.5mm}
\end{figure}

Recently, great advancement has been made in continually training MLMs with unsupervised sentence-level contrastive learning, aiming at creating more discriminative sentence-level representations \citep{giorgi-etal-2021-declutr,carlsson2021semantic,yan-etal-2021-consert,kim-etal-2021-self,liu2021fast,DBLP:journals/corr/abs-2104-08821}. However, such representations are only evaluated as sentence embeddings
and there is no evidence 
that they will
benefit other well-established NLU tasks. 
We show that these approaches hardly bring any benefit to challenging tasks like SQuAD~\cite{DBLP:conf/emnlp/RajpurkarZLL16,DBLP:conf/acl/RajpurkarJL18}.

In this paper, we argue that the key of obtaining more discriminative and transferrable representations lies in learning contrastive and isotropic token-level representations.
To this end, we propose TaCL~(\textbf{T}oken-\textbf{a}ware \textbf{C}ontrastive \textbf{L}earning), a new continual pre-training approach that encourages BERT to learn discriminative token representations. Specifically, our approach involves two models 
(a student and a teacher) that are both initialized from the same pre-trained BERT. During the learning stage, we freeze the parameters of the teacher and continually optimize the student model with (1) the original BERT pre-training objectives (masked language modelling and next sentence prediction) and (2) a newly proposed TaCL objective. The TaCL loss is obtained by contrasting the student representations of masked tokens against the ``reference'' representations produced by the teacher without masking the input tokens. In Figure \ref{fig:overview}, we provide an overview of our approach.

We extensively test our approach on a wide range of English and Chinese benchmarks and illustrate that TaCL brings notable performance improvements on most evaluated datasets (\cref{sec:main_result}). These results validate that more discriminative and isotropic token representations lead to better model performances. Additionally, we highlight the benefits of using our token-level method compared to current state-of-the-art sentence-level contrastive learning techniques on NLU tasks (\cref{sec:ablation_study}). We further analyze the inner workings of TaCL and its impact on the token representation space (\cref{sec:self}). 

Our work, to the best of our knowledge, is the first effort on applying contrastive learning to improve token representations of Transformer models. We hope the findings of this work could facilitate further development of methods on the intersection of contrastive learning and representation learning at a more fine-grained granularities.


\section{Token-aware Contrastive Learning}
Our approach contains two models, i.e., a student $S$ and a teacher $T$, both of which are initialized from the same pre-trained BERT. During learning, we freeze $T$ and only optimize the parameters of $S$. Given an input sequence $x = [x_1, ..., x_n]$, we randomly mask $x$ with the same procedure as in \citet{DBLP:conf/naacl/DevlinCLT19} and feed the masked sequence $\Tilde{x}$ into the student model to produce the contextual representation $\Tilde{h} = [\Tilde{h}_1, ..., \Tilde{h}_n]$. Meanwhile, the teacher model takes the original sequence $x$ as input and produces the representation $h = [h_1, ..., h_n]$ (see Figure~\ref{fig:overview}). The proposed token-aware contrastive learning objective $\mathcal{L}_{\textup{TaCL}}$ is then defined as
\begin{equation}
    \label{eq:cse}
    \setlength{\abovedisplayskip}{3pt}
\setlength{\belowdisplayskip}{3pt}
    -\sum_{i=1}^{n}\mathds{1}(\Tilde{x}_i)\log\frac{\exp(\textup{sim}(\Tilde{h}_i, h_i)/\tau)}{\sum_{j=1}^{n}\exp(\textup{sim}(\Tilde{h}_i,h_j)/\tau)},
\end{equation}
where $\mathds{1}(\Tilde{x}_i)=1$ if $\Tilde{x}_i$ is a masked token, otherwise $\mathds{1}(\Tilde{x}_i)=0$. $\tau$ is a temperature hyper-parameter and $\textup{sim}(\cdot, \cdot)$ computes the cosine similarity. Intuitively, the student learns to make the representation of a masked token closer to its ``reference'' representation produced by the teacher and away from other tokens in the same sequence. As a result, the token representations learnt by the student are more discriminative with respect to distinct tokens, therefore better following an isotropic distribution. Similar to \citet{DBLP:conf/naacl/DevlinCLT19}, we also adopt the masked language modelling $\mathcal{L}_{\textup{MLM}}$ and next sentence prediction $\mathcal{L}_{\textup{NSP}}$ objectives. The overall learning objective $\mathcal{L}$ of the student model during the continual pre-training stage is defined as 
\begin{equation}
    \label{eq:tacl_loss}
    \setlength{\abovedisplayskip}{3pt}
\setlength{\belowdisplayskip}{3pt}
    \mathcal{L} = \mathcal{L}_{\textup{TaCL}} + \mathcal{L}_{\textup{MLM}} + \mathcal{L}_{\textup{NSP}}.
\end{equation}
Note that the learning of the student is fully unsupervised and can be realized using the original pre-training corpus. After the learning is completed, we fine-tune the student model on downstream tasks.

\section{Experiment}

\begin{table*}[t]
    \small
	\centering  
	\renewcommand{\arraystretch}{1.2}
	\setlength{\tabcolsep}{6pt}
	\scalebox{0.83}{
	\begin{tabular}{lcccccccccccccc}
		\hlinewd{0.75pt}
		\multirow{16}{*}{\rotatebox[origin=c]{90}{{English Benchmark}}} &\multirow{2}{*}{\textbf{Model}}&\multicolumn{9}{c}{\textbf{GLUE}}&\multicolumn{2}{c}{\textbf{SQuAD 1.1}}&\multicolumn{2}{c}{\textbf{SQuAD 2.0}}\\
		\cmidrule(lr){3-11}
		\cmidrule(lr){12-13}
		\cmidrule(lr){14-15}
	    &&CoLA&SST-2&MPRC&STS-B&QQP&MNLI&QNLI&RTE&Ave.&EM&F1&EM&F1\\
		\cmidrule(lr){2-15}
	    &&\multicolumn{12}{c}{\textit{Base size models}}\\
	    &$\textup{BERT}_{\textup{base}}\parallel$&52.1&93.5&88.9&85.8&71.2&84.6/83.4&90.5&66.4&79.6&80.8&88.5&-&-\\
	    &$\textup{BERT}_{\textup{base}}$ $\ddagger$&52.2&92.4&89.0&86.4&73.2&\textbf{84.6/84.5}&90.3&63.2&79.8&80.9&88.4&73.4&76.8\\
	    &\makecell[r]{\textit{+MT}$\ddagger$}&51.9&\textbf{92.5}&89.3&87.1&75.8&84.2/84.0&90.6&\textbf{64.1}&80.0&81.0&88.5&73.2&76.3\\
	    \cmidrule(lr){2-15}
	    &$\textup{TaCL}_{\textup{base}}$&\textbf{52.4}&92.3&\textbf{90.8}&\textbf{89.0}&\textbf{80.7}&84.4/84.3&\textbf{91.1}&62.8&\textbf{81.2}&\textbf{81.6}&\textbf{89.0}&\textbf{74.4}&\textbf{77.5}\\
	    \cmidrule(lr){2-15}
	    &&\multicolumn{12}{c}{\textit{Large size models}}\\
	    &$\textup{BERT}_{\textup{large}}\parallel$&60.5&94.9&89.3&86.5&72.1&86.7/85.9&92.7&70.1&82.1&84.1&90.9&78.7&81.9\\
	    &$\textup{BERT}_{\textup{large}}$ $\ddagger$&61.6&93.6&90.2&89.0&81.8&86.4/86.1&\textbf{92.6}&67.2&83.6&84.0&90.8&77.9&81.0\\
	    &\makecell[r]{\textit{+MT}}$\ddagger$&\textbf{62.0}&93.8&90.5&89.1&\textbf{82.5}&86.3/\textbf{86.3}&92.2&66.5&83.7&83.9&90.9&77.8&80.7\\
	    \cmidrule(lr){2-15}
	    &$\textup{TaCL}_{\textup{large}}$&61.1&\textbf{94.1}&\textbf{92.0}&\textbf{89.7}&\textbf{82.5}&\textbf{86.5}/85.9&92.4&\textbf{70.5}&\textbf{84.7}&\textbf{84.2}&\textbf{91.1}&\textbf{78.7}&\textbf{81.9}\\
		\hlinewd{1.5pt}
	\end{tabular}}
	\scalebox{0.87}{
	\begin{tabular}{lcccccccccccc}
        \multirow{9}{*}{\rotatebox[origin=c]{90}{{Chinese Benchmark}}} &\multirow{2}{*}{\textbf{Model}}&\multicolumn{2}{c}{\textbf{Ontonotes}}&\multicolumn{2}{c}{\textbf{MSRA}}&\multicolumn{2}{c}{\textbf{Resume}}&\multicolumn{2}{c}{\textbf{Weibo}}&\textbf{PKU}&\textbf{CityU}&\textbf{AS}\\
        \cmidrule(lr){3-4}
        \cmidrule(lr){5-6}
        \cmidrule(lr){7-8}
        \cmidrule(lr){9-10}
        \cmidrule(lr){11-11}
        \cmidrule(lr){12-12}
        \cmidrule(lr){13-13}
        &&Dev&Test&Dev&Test&Dev&Test&Dev&Test&Test&Test&Test\\
        \cmidrule(lr){2-13}
	    &&\multicolumn{10}{c}{${\spadesuit}$ and ${\diamondsuit}$ published in \citet{DBLP:conf/acl/LiYQH20} and \citet{DBLP:conf/nips/MengWWLNYLHSL19}}\\
		&$\textup{BERT}_{\textup{base}}$&-&80.14$^{\spadesuit}$&-&94.95$^{\spadesuit}$&-&95.53$^{\spadesuit}$&-&68.20$^{\spadesuit}$&96.50$^{\diamondsuit}$&97.60$^{\diamondsuit}$&96.50$^{\diamondsuit}$\\
		\cmidrule(lr){2-13}
		&$\textup{BERT}_{\textup{base}}\ddagger$&78.29&80.23&94.13&94.97&95.37&95.70&70.63&67.98&96.51&97.83&96.58\\
		&\makecell[r]{\textit{+MT}}$\ddagger$&78.42&80.36&94.20&95.01&95.29&95.62&70.81&68.02&96.53&97.79&96.54\\
		\cmidrule(lr){2-13}
		&$\textup{TaCL}_{\textup{base}}$&\textbf{79.73}&\textbf{82.42}&\textbf{94.58}&\textbf{95.44}&\textbf{96.23}&\textbf{96.45}&\textbf{72.32}&\textbf{69.54}&\textbf{96.75}&\textbf{98.18}&\textbf{96.75}\\
		\hlinewd{0.75pt}
	\end{tabular}}
    \caption{Benchmark Results. $\parallel$: published in \citet{DBLP:conf/naacl/DevlinCLT19}; and $\ddagger$: models from our implementations.}
	\label{tb:benchmark}
	\vspace{-1.5mm}
\end{table*}

We test our approach on a wide range of benchmarks in two languages. For English benchmarks, we evaluate the $\textup{BERT}_{\textup{base}}$ and $\textup{BERT}_{\textup{large}}$ models. For Chinese benchmarks, we test the $\textup{BERT}_{\textup{base}}$  model.\footnote{All models are officially released by \citet{DBLP:conf/naacl/DevlinCLT19}.}
After initializing the student and teacher, we continually pre-train the student on the same Wikipedia corpus as in \citet{DBLP:conf/naacl/DevlinCLT19} for 150k steps. The training samples are truncated with a maximum length of 256 and the batch size is set as 256. The temperature $\tau$ in Eq.~(\ref{eq:cse}) is set as 0.01. Same as \citet{DBLP:conf/naacl/DevlinCLT19}, we optimize the model with Adam optimizer \cite{DBLP:journals/corr/KingmaB14} with weighted decay, and an initial learning rate of 1e-4 (with warm-up ratio of 10\%).


\subsection{Evaluation Benchmarks}
%
For English benchmarks, we use the GLUE dataset \cite{DBLP:conf/iclr/WangSMHLB19} which contains a variety of sentence-level classification tasks covering textual entailment (RTE and MNLI), question-answer entailment (QNLI), paraphrase (MRPC), question paraphrase (QQP), textual similarity (STS-B), sentiment (SST-2), and linguistic acceptability (CoLA). Our evaluation metrics are Spearman correlation for STS-B, Matthews correlation for CoLA, and accuracy for the other tasks; the macro average score is also reported. Additionally, we conduct experiments on SQuAD 1.1 \cite{DBLP:conf/emnlp/RajpurkarZLL16} and 2.0 \cite{DBLP:conf/acl/RajpurkarJL18} datasets that evaluate the model's performance on the token-level answer-extraction task. The dev set results of Exact-Match (EM) and F1 scores are reported.


For Chinese benchmarks, we evaluate our model on two token-level labelling tasks, including name entity recognition (NER) and Chinese word segmentation (CWS). For NER, we use the Ontonotes \cite{OntoNotes4}, MSRA \cite{DBLP:conf/acl-sighan/Levow06}, Resume \cite{DBLP:conf/acl/ZhangY18}, and Weibo \cite{DBLP:conf/eacl/SunH17} datasets. For CWS, we use the PKU, CityU, and AS datasets from SIGHAN 2005 \cite{DBLP:conf/acl-sighan/Emerson05} for evaluation. The standard F1 score is used for evaluation.

\noindent\textbf{Baselines:} We compare against two baselines: (1) the original BERT used to initialize the student and teacher; (2) BERT\textit{+MT} (BERT with more training) which is acquired by continually pre-training the original BERT on Wikipedia for 150k steps\footnote{The number of steps is set  the same as our TaCL training.} using the original BERT pre-training objectives.

\subsubsection{Benchmark Results}
\label{sec:main_result}
Table \ref{tb:benchmark} reports the results on English and Chinese benchmarks.\footnote{For all tasks, the average results over five runs are reported.} 
We observe that, on most sequence-level classification tasks in GLUE, TaCL outperforms BERT and BERT\textit{+MT}. 
Additionally, on all token-level benchmarks (SQuAD, NER, and CWS), TaCL consistently and notably surpasses other baselines. These results indicate that the learning of an isotropic token representation space is beneficial for the model's performance, especially on the token-centric tasks.

\begin{table}[t]
    \small
	\centering  
	\renewcommand{\arraystretch}{1.2}
	\setlength{\tabcolsep}{6pt}
	\scalebox{0.82}{
	\begin{tabular}{ccccc}
		\hlinewd{0.75pt}
		\textbf{Model}&\textbf{$\mathcal{L_{\textup{MLM}}}$ + $\mathcal{L_{\textup{NSP}}}$}&\textbf{CL}&\textbf{SQuAD 1.1}&\textbf{SQuAD 2.0}\\
		\hline
		BERT&\checkmark&$\times$&80.8/88.5&73.4/76.8\\
		\hline
		\multicolumn{5}{c}{\textit{Sentence-Level Contrastive Methods}}\\
		DeCLUTR&$\times$&Sen.&79.9/87.6&72.1/75.4\\
		SimCSE&$\times$&Sen.&80.2/88.0&72.5/75.7\\
		MirrorBERT&$\times$&Sen.&80.3/88.1&72.7/75.9\\
		\hline
		\multicolumn{5}{c}{\textit{Ablated Models}}\\
		model-1&\checkmark&Sen.&80.5/88.3&73.1/76.5\\
		model-2&$\times$&Tok.&81.3/88.7&73.8/77.1\\
		\hline
		TaCL&\checkmark&Tok.&\textbf{81.6/89.0}&\textbf{74.4/77.5}\\
		\hlinewd{0.75pt}
	\end{tabular}}
    \caption{Comparison of various sentence- and token-level contrastive learning methods. ``Sen.'' or ``Tok.'' denotes training with sentence- or token-level contrastive objectives. Scores of (EM/F1) are reported.}
    	\vspace{-1.5mm}
	\label{tb:ablation_study}
\end{table}



\subsection{Analysis}
In this section, we present further comparisons and in-depth analysis of the proposed approach. 

\subsubsection{Sentence-Level vs. Token-Level CL}
\label{sec:ablation_study}
We compare TaCL against existing sentence-level contrastive learning methods, including DeCLUTR~\cite{giorgi-etal-2021-declutr}, SimCSE~\cite{DBLP:journals/corr/abs-2104-08821}, and MirrorBERT~\cite{liu2021fast}. We also include two ablated models to study the effect of different combinations of pre-training objectives. Specifically, the ablated model-1 is initialized with BERT and trained with the original BERT objectives ($\mathcal{L}_{\textup{MLM}}$ and $\mathcal{L}_{\textup{NSP}}$) \textbf{plus} the sentence-level contrastive objective as proposed in \citet{liu2021fast}. The ablated model-2 is initialized with BERT and trained \textbf{only} with the proposed token-aware contrastive objective of Eq. (\ref{eq:cse}). Note that all compared models have the same size as the  $\textup{BERT}_{\textup{base}}$ model.


Table \ref{tb:ablation_study} shows the performance of different models on SQuAD. We observe decreased performance of existing sentence-level contrastive methods 
compared with the original BERT. This could be attributed to the fact that such methods only focus on learning sentence-level representations while ignoring the learning of individual tokens. This behaviour is undesired for tasks like SQuAD that demands informative token representations. Nonetheless, the ablated model-1 shows that the original BERT pre-training objective ($\mathcal{L}_{\textup{MLM}}$ and $\mathcal{L}_{\textup{NSP}}$) remedies, to some extent, the performance degeneration caused by the sentence-level contrastive methods. On the other hand, the ablated model-2 demonstrates that our token-aware contrastive objective helps the model to achieve improved results by learning better token representations. 

\begin{figure}[tb] 
  \centering    
  \setlength{\abovecaptionskip}{3pt}
  \includegraphics[width=0.46\textwidth]{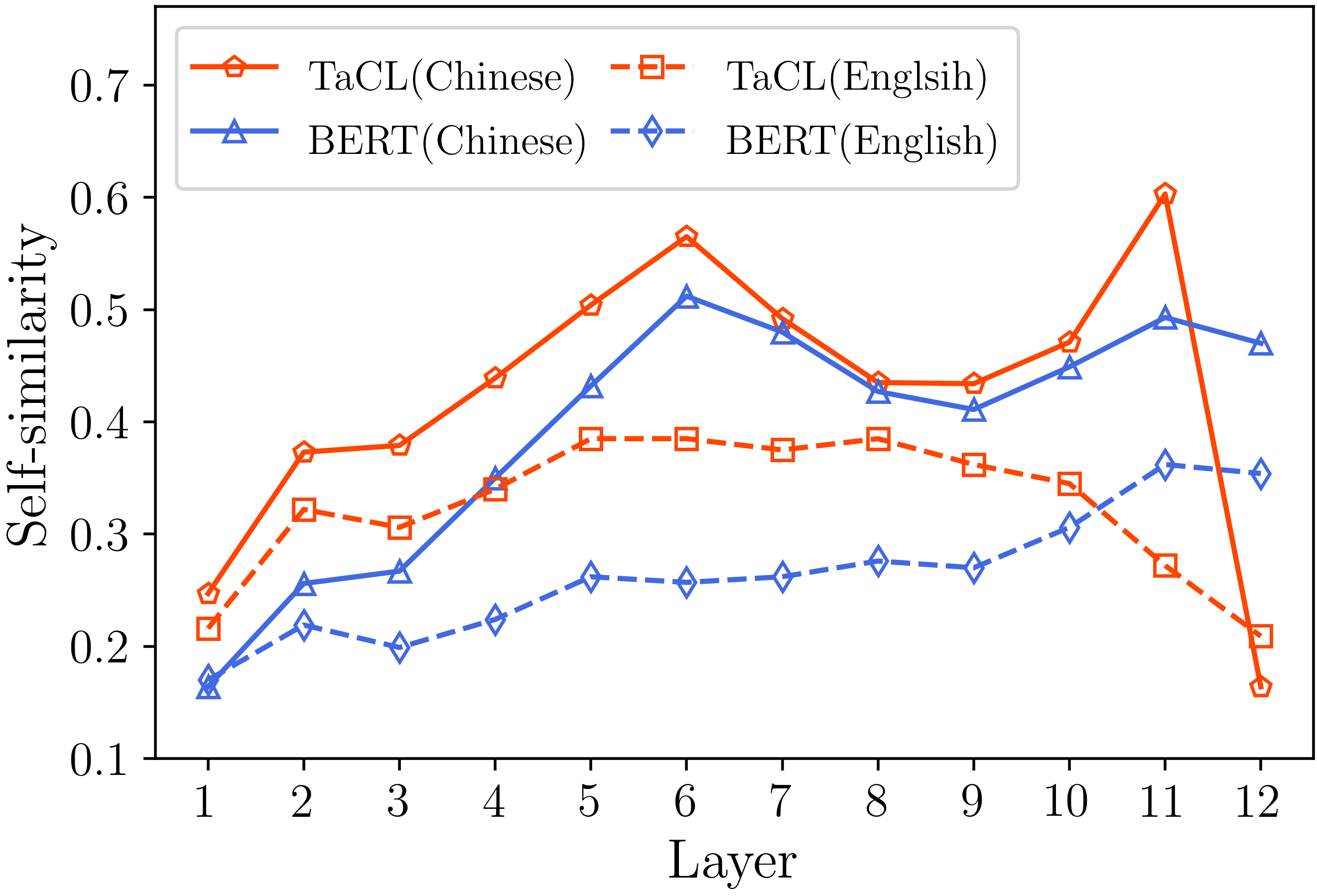}
  \caption{Layer-wise representation self-similarity.}
  \label{fig:cross_similarity}
  \vspace{-1.5mm}
\end{figure}

\subsubsection{Token Representation Self-similarity}
\label{sec:self}
\label{sec:cross_similarity_analysis}
To analyze the token representations learnt by TaCL and BERT, we follow \citet{DBLP:conf/emnlp/Ethayarajh19} and define the averaged self-similarity of the token representations within one sequence $x=[x_1, ..., x_n]$ as,
\begin{equation}
\setlength{\abovedisplayskip}{3pt}
\setlength{\belowdisplayskip}{3pt}
    s(x) = \frac{1}{n(n-1)}\sum_{i=1}^{n}\sum_{j=1,j\neq i}^{n}\textup{cosine}(h_i, h_j),
\end{equation}
where $h_i$ and $h_j$ are the token representations of $x_i$ and $x_j$ produced by the model. Intuitively, a lower $s(x)$ indicates that the representations of tokens within the sequence $x$ are less similar to each other, therefore being more discriminative.

We sample 50k sentences from both Chinese and English Wikipedia and compute the self-similarity of representations over different layers. 
Figure~\ref{fig:cross_similarity} plots the results of $\textup{TaCL}_{\textup{base}}$ and $\textup{BERT}_{\textup{base}}$ averaged over all sentences. We see that, in the intermediate layers, the self-similarity of TaCL is higher than BERT's. 
In contrast, at the top layer (layer 12), TaCL's self-similarity becomes notably lower than BERT's, demonstrating that the final output token representations of TaCL are more discriminative. 

\begin{figure}[tb] 
  \centering    
  \setlength{\abovecaptionskip}{3pt}
  \includegraphics[width=0.48\textwidth]{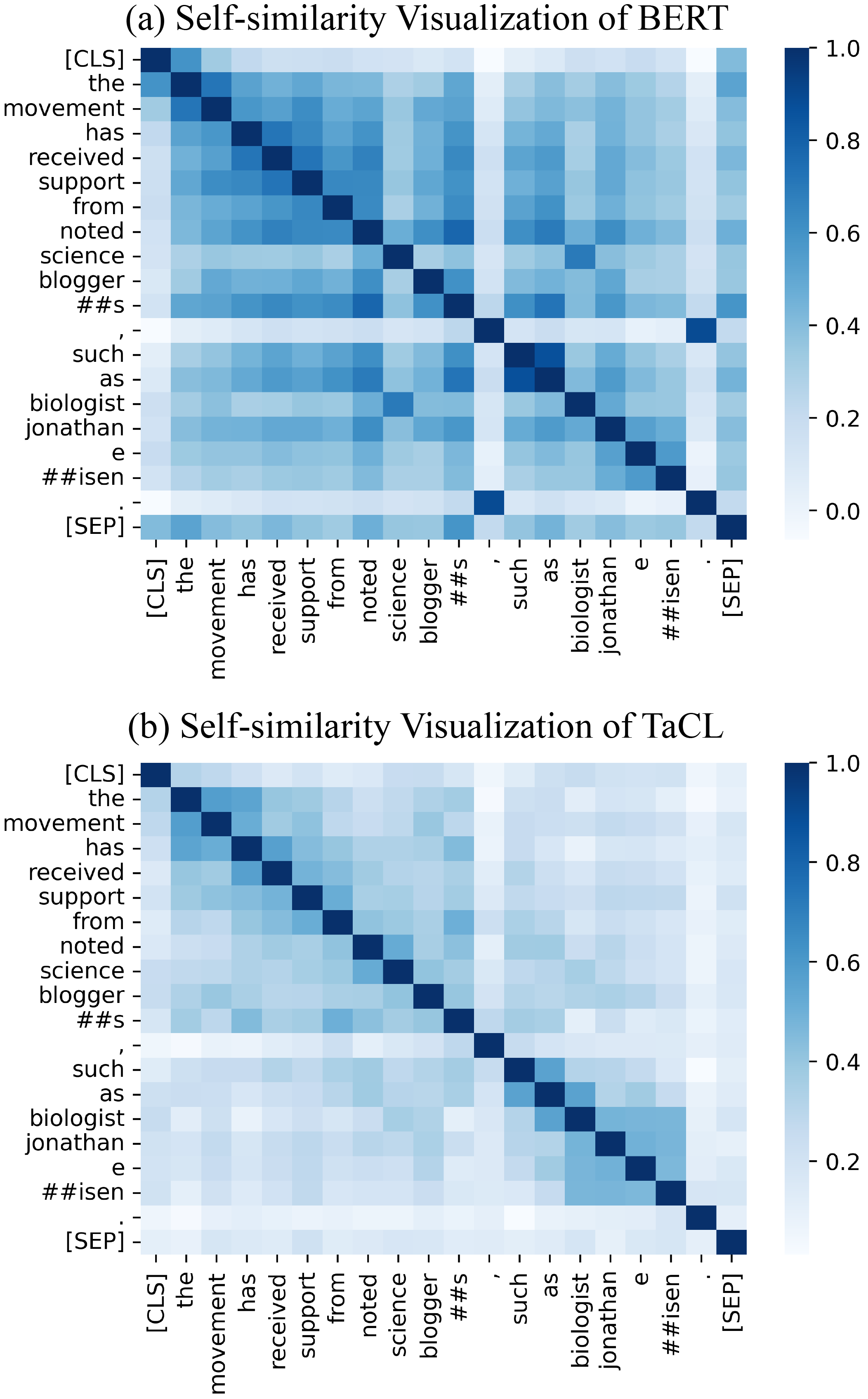}
  \caption{Self-similarity Matrix Visualization: (a) BERT and (b) TaCL. (best viewed in color)}
  \label{fig:heatmap}
\end{figure}

\paragraph{Qualitative Analysis.} 
%
We sample one sentence from Wikipedia and visualize the self-similarity matrix $M$ (where $M_{i,j}=\textup{cosine}(h_i, h_j)$) produced by $\textup{BERT}_{\textup{base}}$ and $\textup{TaCL}_{\textup{base}}$. The results are shown in Figure \ref{fig:heatmap}, where a darker color denotes a higher self-similarity score.\footnote{The entries $M_{i,i}$ in the diagonal have a $1.0$ self-similarity by definition, as $\textup{cosine}(h_i,h_i)=1.0$.} We see that, as compared with BERT (Fig. \ref{fig:heatmap}(a)), the self-similarities of TaCL (Fig. \ref{fig:heatmap}(b)) are much lower in the off-diagonal entries. This further highlights that the individual token representations of TaCL are more discriminative, which in return leads to improved model performances as demonstrated (\cref{sec:main_result}, \cref{sec:ablation_study}).

\section{Conclusion}
In this work, we proposed TaCL, a novel approach that applies token-aware contrastive learning for the continual pre-training of BERT. Extensive experiments are conducted on a wide range of English and Chinese benchmarks. The results show that our approach leads to notable performance improvement across all evaluated benchmarks. We then delve into the inner-working of TaCL and demonstrate that our performance gain comes from a more discriminative distribution of token representations. 

\section*{Acknowledgments}
The first author would like to thank Jialu Xu and Piji Li for the insightful discussions and support. Many thanks to our anonymous reviewers and area chairs for their suggestions and comments.

\section*{Ethical Statement}
We honor and support the ACL code of Ethics. Language model pre-training aims to improve the system's performance on downstream NLU tasks. All pre-training corpora and pre-trained models used in this work are publically available. In addition, all evaluated datasets are from previously published works, and in our view, do not have any attached privacy or ethical issues.

\bibliographystyle{acl_natbib}
\bibliography{anthology,acl2021}

\clearpage

\appendix

\section{Statistics of Evaluated Benchmarks}
\subsection{English Benchmarks}
\begin{table}[h]
    \small
	\centering  
	\renewcommand{\arraystretch}{1.2}
	\setlength{\tabcolsep}{6pt}
	\scalebox{0.88}{
	\begin{tabular}{cccc}
		\hlinewd{0.75pt}
		\textbf{Dataset}&\textbf{Train}&\textbf{Test}&\textbf{Evaluation Metric}\\
		\hline
		CoLA&8.5k&1k&Matthews correlation\\
		SST-2&67k&1.8k&accuracy\\
		MRPC&3.7k&1.7K&accuracy\\
		STS-B&7k&1.4k&Spearman correlation\\
		QQP&364k&391k&accuracy\\
		MNLI&393k&20k&matched/mismatched accuracy\\
		QNLI&105k&5.4k&accuracy\\
		RTE&2.5k&3k&accuracy\\
		\hlinewd{0.75pt}
	\end{tabular}}
    \caption{GLUE Statistics}
    	\vspace{-1.5mm}
	\label{tb:glue_stat}
\end{table}

\begin{table}[h]
    \small
	\centering  
	\renewcommand{\arraystretch}{1.2}
	\setlength{\tabcolsep}{6pt}
	\scalebox{0.95}{
	\begin{tabular}{cccc}
		\hlinewd{0.75pt}
		\textbf{Dataset}&\textbf{Train}&\textbf{Dev}&\textbf{Evaluation Metric}\\
		\hline
		1.1&87.6k&10.6k&Exact-Match/F1\\
		2.0&130.3k&11.9k&Exact-Match/F1\\
		\hlinewd{0.75pt}
	\end{tabular}}
    \caption{SQuAD Statistics}
    	\vspace{-1.5mm}
	\label{tb:squad_stat}
\end{table}

\subsection{Chinese Benchmarks}
\begin{table}[h]
    \small
	\centering  
	\renewcommand{\arraystretch}{1.2}
	\setlength{\tabcolsep}{6pt}
	\scalebox{0.9}{
	\begin{tabular}{ccccc}
		\hlinewd{0.75pt}
		\textbf{Dataset}&\textbf{Train}&\textbf{Dev}&\textbf{Test}&\textbf{Evaluation Metric}\\
		\hline
		Ontonotes&15.7k&4.3k&4.3k&F1\\
		MSRA&37.0k&9.3k&4.4k&F1\\
		Resume&3.8k&0.5k&0.5k&F1\\
		Weibo&1.4k&0.3k&0.3k&F1\\
		\hlinewd{0.75pt}
	\end{tabular}}
    \caption{NER Dataset Statistics}
    	\vspace{-1.5mm}
	\label{tb:ner_stat}
\end{table}

\begin{table}[h]
    \small
	\centering  
	\renewcommand{\arraystretch}{1.2}
	\setlength{\tabcolsep}{6pt}
	\scalebox{0.9}{
	\begin{tabular}{cccc}
		\hlinewd{0.75pt}
		\textbf{Dataset}&\textbf{Train}&\textbf{Test}&\textbf{Evaluation Metric}\\
		\hline
		PKU&19.1k&1.9k&F1\\
		CityU&53.0k&1.5k&F1\\
		AS&708.9k&14.4k&F1\\
		\hlinewd{0.75pt}
	\end{tabular}}
    \caption{CWS Dataset Statistics}
    	\vspace{-1.5mm}
	\label{tb:cws_stat}
\end{table}

\section{Related Work}
\paragraph{Pre-trained Language Models.} Since the introduction of BERT \cite{DBLP:conf/naacl/DevlinCLT19}, the research community has witnessed remarkable progress in the field of language model pre-training on a large amount of free text. Such advancements have led to significant progresses in a wide range of natural language understanding (NLU) tasks \cite{DBLP:journals/corr/abs-1907-11692,DBLP:conf/nips/YangDYCSL19,DBLP:conf/iclr/ClarkLLM20,DBLP:journals/corr/abs-2110-06612} and text generation tasks \cite{radford2019language,DBLP:conf/acl/LewisLGGMLSZ20,DBLP:journals/jmlr/RaffelSRLNMZLL20,DBLP:conf/eacl/SuCWVBLC21,DBLP:conf/acl/SuVBWC21,DBLP:journals/taslp/SuWCBKC21,DBLP:journals/corr/abs-2109-14739,DBLP:journals/corr/abs-2108-13740,DBLP:journals/corr/abs-2108-12516,DBLP:journals/corr/abs-2109-02492}


\paragraph{Contrastive Learning.} Generally, contrastive learning methods distinguish observed data points from fictitious negative samples. They have been widely applied to various computer vision areas, including image \cite{DBLP:conf/cvpr/ChopraHL05,oord2018representation} and video \cite{DBLP:conf/iccv/WangG15,DBLP:conf/icra/SermanetLCHJSLB18}. Recently, \citet{chen2020simple} proposed a simple framework for contrastive learning of visual representations (SimCLR) based on multi-class N-pair loss. \citet{radford2021learning,pmlr-v139-jia21b} applied the contrastive learning approach for language-image pretraining. \citet{xu2021videoclip, yang2021taco} proposed a contrastive pre-training approach for video-text alignment.

In the field of NLP, numerous approaches have been proposed to learn better sentence-level \cite{reimers-gurevych-2019-sentence,wu2020clear,meng2021coco,liu2021fast,DBLP:journals/corr/abs-2104-08821,DBLP:conf/acl/Su0ZLBC0C020} and lexical-level \citep{liu-etal-2021-self,vulic-etal-2021-lexfit,liu-etal-2021-mirrorwic,Wang2021PhraseBERTIP} representations using contrastive learning. Different from our work, none of these studies specifically investigates how to utilize contrastive learning for improving general-purpose token-level representations. Beyond representation learning, contrastive learning has also been applied to other NLP applications such as NER \citep{das2021container} and summarisation \citep{liu-liu-2021-simcls}, knowledge probing for pre-trained language models \citep{DBLP:journals/corr/abs-2110-08173}, and open-ended text generation \citep{DBLP:journals/corr/abs-2202-06417}.

\paragraph{Continual Pre-training.} Many researchers \cite{xu-etal-2019-bert, gururangan2020don, pan2021multilingual} have investigated how to continually pre-train the model to alleviate the task- and domain-discrepancy between the pre-trained models and the specific target task. In contrast, our proposed approach studies how to apply continual pre-training to directly improve the quality of model representations which is transferable and beneficial to a wide range of benchmark tasks.

\begin{figure}[tb] 
  \centering    
  \setlength{\abovecaptionskip}{3pt}
  \includegraphics[width=0.5\textwidth]{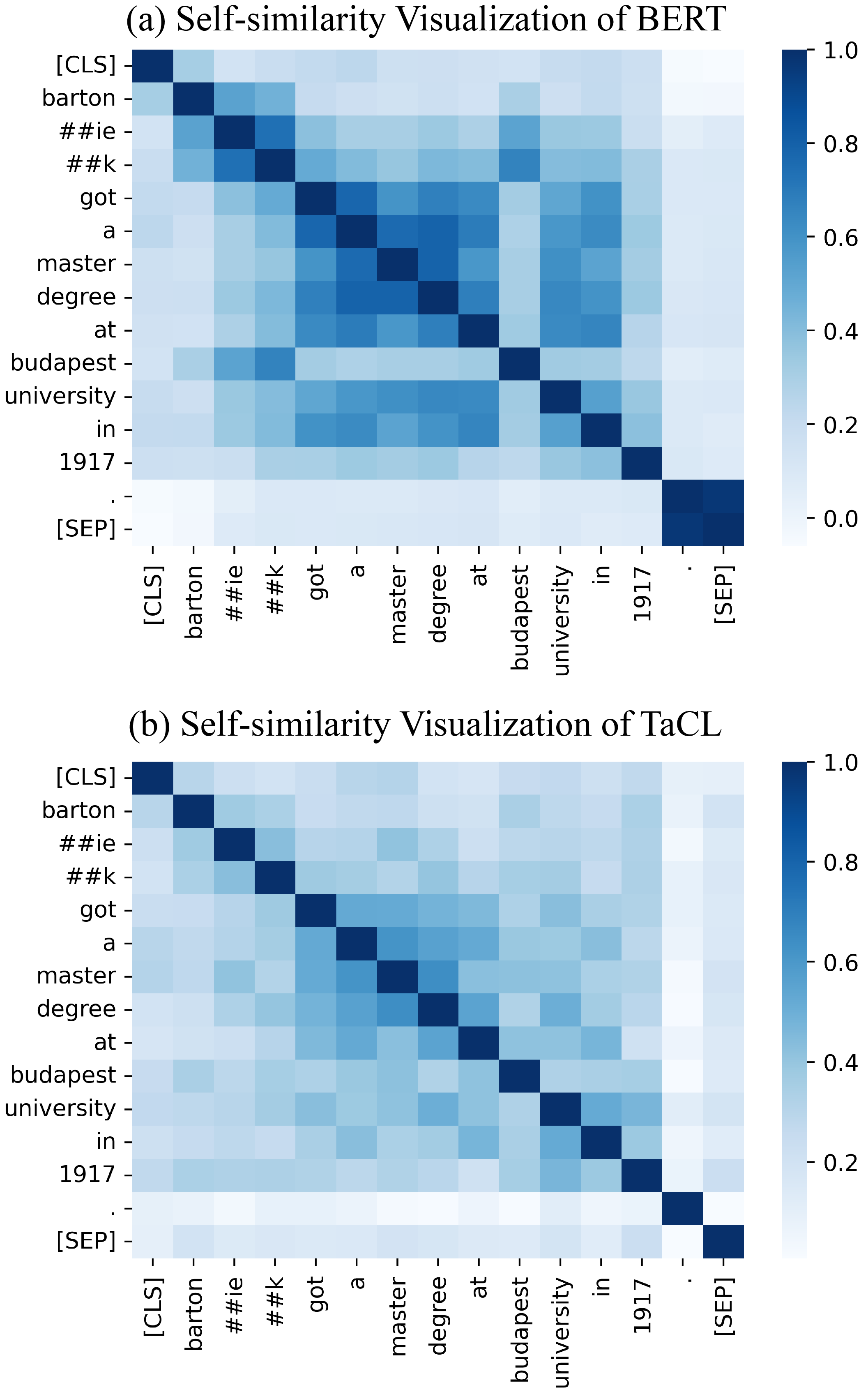}
  \caption{\textbf{Example 2}: self-similarity matrix visualization of (a) BERT and (b) TaCL. (best viewed in color)}
  \label{fig:heatmap_2}
\end{figure}

\begin{figure}[h] 
  \centering    
  \setlength{\abovecaptionskip}{3pt}
  \includegraphics[width=0.5\textwidth]{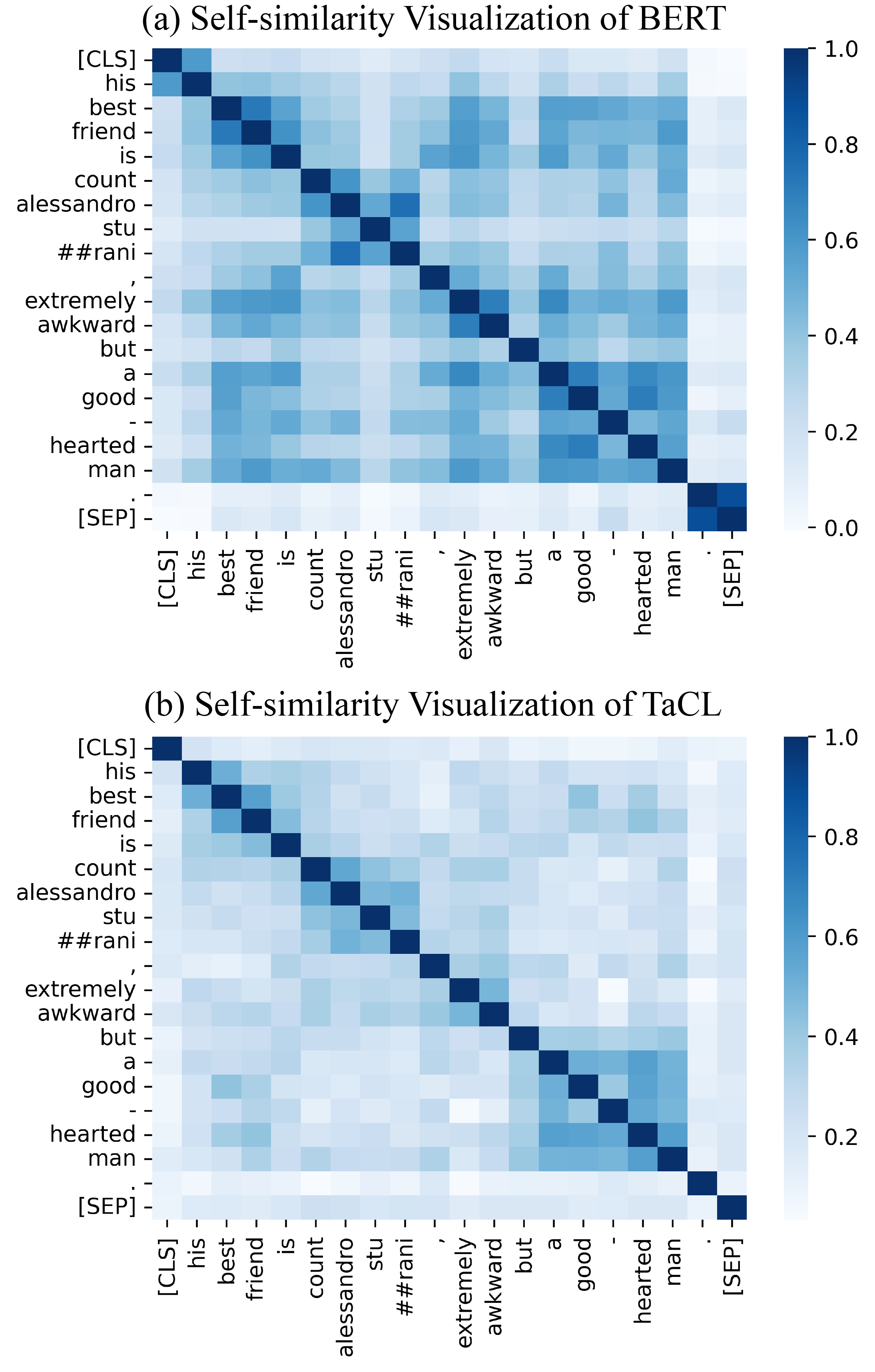}
  \caption{\textbf{Example 3}: self-similarity matrix visualization of (a) BERT and (b) TaCL. (best viewed in color)}
  \label{fig:heatmap_3}
\end{figure}

\begin{figure}[t] 
  \centering    
  \setlength{\abovecaptionskip}{3pt}
  \includegraphics[width=0.5\textwidth]{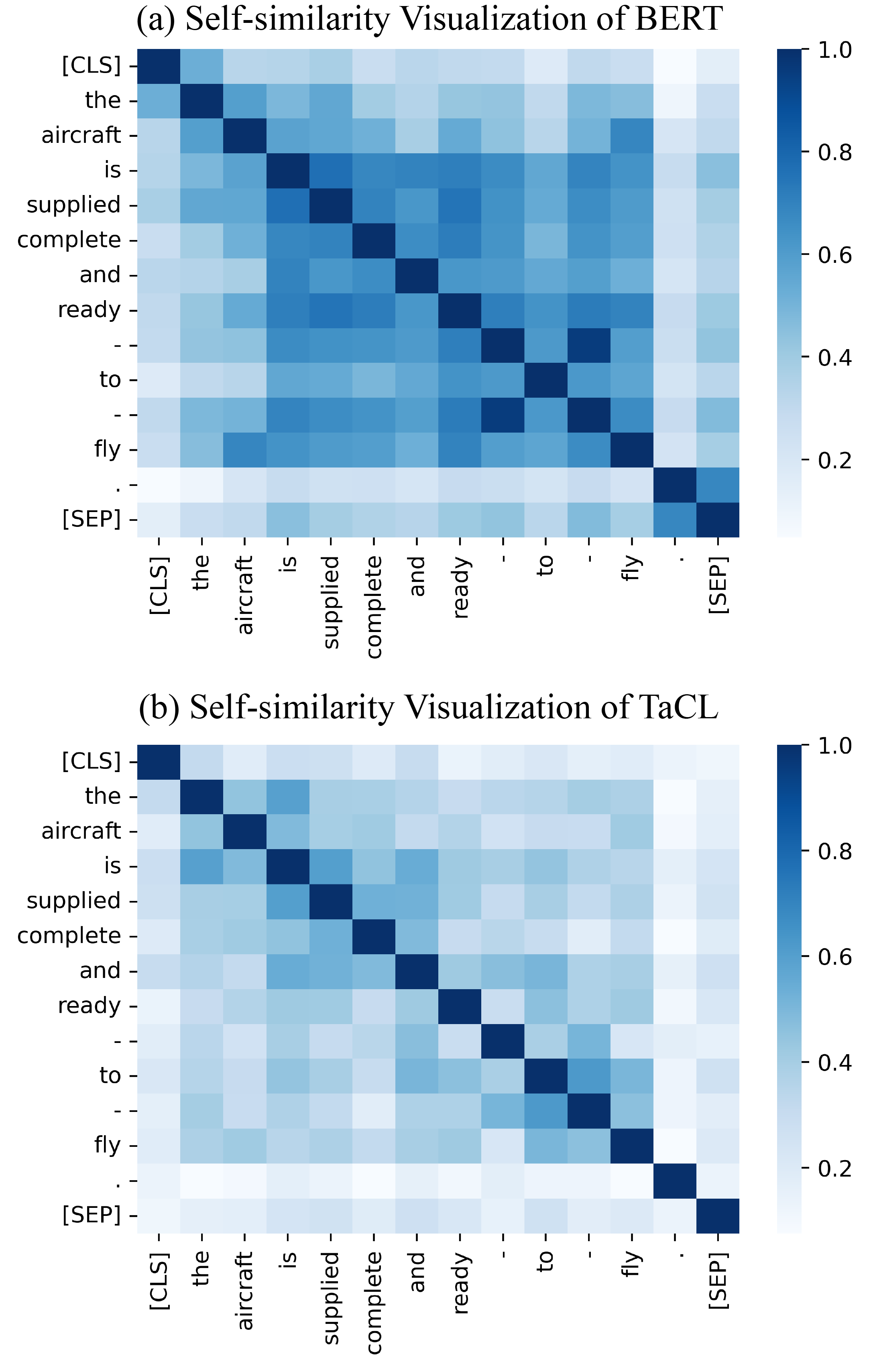}
  \caption{\textbf{Example 4}: self-similarity matrix visualization of (a) BERT and (b) TaCL. (best viewed in color)}
  \label{fig:heatmap_4}
\end{figure}

\section{More Self-similarity Visualizations}
In Figure \ref{fig:heatmap_2}, \ref{fig:heatmap_3}, and \ref{fig:heatmap_4}, we provide three more comparisons between the self-similarity matrix produced by TaCL and BERT (the example sentences are randomly sampled from Wikipedia).\footnote{All results are generated by models with base size.} From the figures, we can draw the same conclusion as in section \cref{sec:self}, that the token representations of BERT follow an anisotropic distribution and are less discriminative. On the other hand, the token representations of TaCL better follow an isotropic distribution, therefore different tokens become more distinguishable with respect to each other.

\end{document}